\pdfoutput=1

\documentclass[11pt]{article}

\usepackage[preprint]{acl}

\usepackage{times}
\usepackage{latexsym}
\usepackage[most]{tcolorbox}
\usepackage[T1]{fontenc}

\usepackage[utf8]{inputenc}
\usepackage{graphicx}
\usepackage{multirow}
\usepackage{booktabs}
\usepackage{enumitem}
\usepackage{amssymb}
\usepackage{pifont}

\usepackage{microtype}

\usepackage{inconsolata}

\usepackage{graphicx}

\title{Lemma Dilemma: On Lemma Generation Without Domain- or Language-Specific Training Data}

\author{
 \textbf{Olia Toporkov\textsuperscript{1}},
 \textbf{Alan Akbik\textsuperscript{2}},
 \textbf{Rodrigo Agerri\textsuperscript{1}}
 \\
  \textsuperscript{1}HiTZ Center - Ixa, University of the Basque Country UPV/EHU
  \\
  \textsuperscript{2}Humboldt-Universität zu Berlin
  \\
   \small{
  \texttt{\{olia.toporkov, rodrigo.agerri\}@ehu.eus}, 
  \texttt{alan.akbik@hu-berlin.de}}
}

\begin{document}
\maketitle
\begin{abstract}
Lemmatization is the task of transforming all words in a given text to their dictionary forms. 
While large language models (LLMs) have demonstrated their ability to achieve competitive results across a wide range of NLP tasks, there is no prior evidence of how effective they are in the contextual lemmatization task. In this paper, we empirically investigate the capacity of the latest generation of LLMs to perform in-context lemmatization, comparing it to the traditional fully supervised approach. In particular, we 
consider the setting in which supervised training data is not available for a target domain or language, comparing (i) encoder-only supervised approaches, fine-tuned out-of-domain, and (ii) cross-lingual methods, against direct in-context lemma generation with LLMs. Our experimental investigation across 12 languages of different morphological complexity finds that, while encoders remain competitive in out-of-domain settings when fine-tuned on gold data, current LLMs reach state-of-the-art results for most languages by directly generating lemmas in-context without prior fine-tuning, provided just with a few examples. Data and code available upon publication: \url{https://github.com/oltoporkov/lemma-dilemma}


\end{abstract}

\section{Introduction}

Lemmatization is one of the core NLP tasks widely used during data pre-processing in such areas as information extraction, named entity recognition, and sentiment analysis, and is of particular importance for languages with complex morphology. To lemmatize a word means to transform its inflected form (e.g. \textit{chose, chosen}) into its dictionary-based form, also known as lemma (e.g. \textit{choose}), according to the definition of the contextual lemmatization task in SIGMORPHON 2019 \cite{mccarthy-etal-2019-sigmorphon}.

\begin{figure}[h] %
\centering
\includegraphics[width=0.5\textwidth]{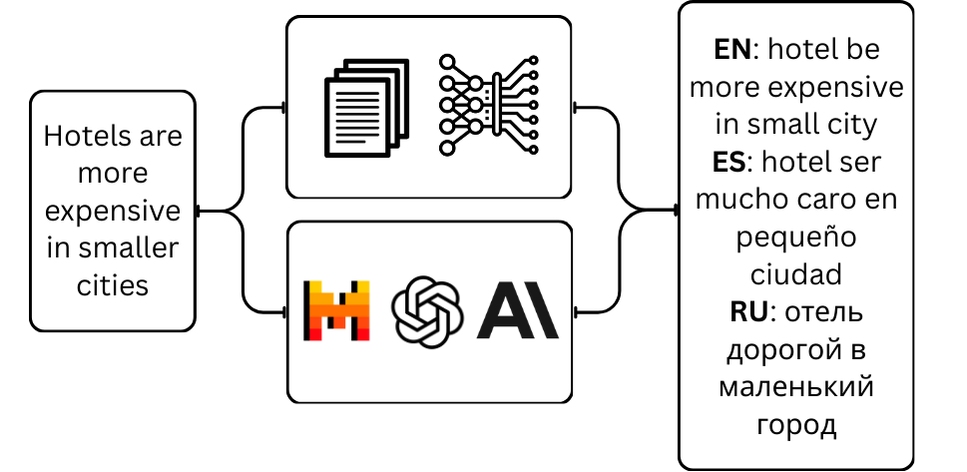}
\caption{General overview of the lemmatization task process for in-context learning and supervised approaches.}
\label{fig:example_lemmatization}
\end{figure}

Most recent approaches tend to address contextual lemmatization as a supervised classification approach, which was first proposed by \citet{chrupala-etal-2008-learning} and became the core idea in the architecture of a variety of contextual lemmatizers \citep{malaviya-etal-2019-simple, straka-etal-2019-udpipe, yildiz-tantug-2019-morpheus}. This method learns to determine the minimum number of edits necessary to convert the word into its lemma. Such techniques require a large amount of annotated data, which can be especially challenging for languages with rich morphology \cite{straka-etal-2019-udpipe, yildiz-tantug-2019-morpheus}. Apart from that, the majority of lemmatization systems are evaluated mostly in-domain, while their real application is almost always out-of-domain, namely, out of the scope of the data they have been trained on. Previous work has demonstrated that lemmatizers' performance worsens substantially when deployed out-of-domain \cite{toporkov-agerri-2024-role}.

The recent generation of large language models (LLMs) has exhibited strong capabilities on a wide variety of NLP tasks, such as reasoning, problem-solving, code generation, information extraction, and text composition \cite{NEURIPS2020_1457c0d6, clark-etal-2021-thats, wang-etal-2022-super, bubeck2023-gpt4, sviridova-etal-2024-casimedicos, sainz-2024-gollie}. However, as it has also been claimed, when LLMs are evaluated on languages other than English or other high-resource languages (e.g., Spanish), their performance is not as good as expected \cite{blasi-etal-2022-systematic, arnett-bergen-2025-language, figueras2025truth}. As for the contextual lemmatization task, to the best of our knowledge, there is no empirical evidence on the capacity of such models in generating the correct lemmas, especially for high-inflected and low-resource languages. 

The scarcity of manually annotated data is another problem for training competitive contextual lemmatizers. The attempts to create large annotated corpora, such as Universal Dependencies \citep{nivre-etal-2017-universal} and the UniMorph project \citep{mccarthy-etal-2020-unimorph}, aim to bridge this resource gap, but they still cover a very limited scope of languages and domains \citep{joshi-etal-2020-state}. In response to data scarcity, model- and data-based cross-lingual transfer for sequence labeling have been proposed \citep{garcia-ferrero-etal-2022-model, chen-etal-2023-frustratingly, yeginbergen-etal-2024-argument}. These approaches focus on overcoming the lack of data in a target language by either fine-tuning a pre-trained multilingual model on a source language (usually English) in order to make predictions for any of the target languages included in the model pre-training (\textit{model-transfer}); or automatically generating labeled data for the target language (\textit{data-transfer}), which is then used to train a sequence labeling model.

Considering the lemmatization challenges mentioned above, in this paper we raise the following research questions (RQ): \textbf{RQ1}: To what extent can the latest generation of large language models directly obtain lemmas in the target language without prior fine-tuning, especially for the languages with complex morphology? \textbf{RQ2}: Could in-context lemma generation, produced by LLMs, be an alternative in solving the lemmatization task out-of-domain? \textbf{RQ3}: What is the best strategy in the scenario where the data in the target language or domain is non-existent or difficult to access? The setup to investigate these RQs is illustrated in Figure \ref{fig:example_lemmatization}.

Hence, our contributions are the following: (i) we empirically investigate the ability of LLMs to perform in-context lemma generation across languages of different morphological complexity; (ii) we address the out-of-domain problem by comparing the performance of LLMs against encoder-only models fine-tuned on different data distribution, and (iii) we conduct a comparative analysis of the methods to overcome the data scarcity in the target language for the lemmatization task, namely, model-transfer, data-transfer and direct lemma generation, using the data in English language as a source. Overall, our results suggest that, while fine-tuning encoders on gold data remains a competitive option for out-of-domain settings, generative LLMs reach state-of-the-art results in lemmatization by directly generating lemmas in-context without prior fine-tuning, provided just with a few examples.

\section{Related Work}

Approaches to perform lemmatization evolved from being rule-based language-dependent systems \citep{karttunen-etal-1992-two, oflazer-1993-two, Alegria1996AutomaticMA, Segalovich2003AFM, jongejan2009automatic} to advanced multilingual architectures, trained on a large amount of annotated data \citep{muller-etal-2015-joint, bergmanis-goldwater-2018-context, malaviya-etal-2019-simple, straka-etal-2019-udpipe}. The idea of addressing lemmatization as a sequence labeling task, where the labels are induced as the minimum amount of edits necessary to convert an inflected word into its lemma, was first proposed by \citet{chrupala-etal-2008-learning} and has been adopted in a wide variety of systems such as the system by \citet{gesmundo-samardzic-2012-lemmatisation}, IXA pipes \citep{agerri-etal-2014-ixa,agerri_rigau}, Lemming \citep{muller-etal-2015-joint}, the system of \citet{malaviya-etal-2019-simple} and Morpheus \citep{yildiz-tantug-2019-morpheus}, among others.

The advancement of supervised techniques involving deep learning algorithms, and the development of the Transformer architecture \citep{vaswani2017attention} and Transformer-based masked language models (MLMs) such as BERT \citep{devlin-etal-2019-bert} and the multilingual XLM-RoBERTa \citep{conneau-etal-2020-unsupervised}, have massively improved the performance of current contextual lemmatizers. Thus, in the SIGMORPHON 2019 Shared Task \citep{mccarthy-etal-2019-sigmorphon} on contextual lemmatization, most of the participant systems were based on MLMs \citep{straka-etal-2019-udpipe, kondratyuk-2019-cross,  shadikhodjaev-lee-2019-cbnu}. 

The evaluation of contextual lemmatizers is almost always performed in-domain, namely, on the same data distribution used during the fine-tuning process. However, in practice, lemmatizers are usually deployed out-of-domain, which results in a significant performance degradation, especially for high-inflected languages \citep{toporkov-agerri-2024-role}.

The rise and constant development of LLMs has demonstrated remarkable abilities in dealing with a wide variety of NLP tasks such as language understanding, reasoning, language generation, code generation, and query response, especially when using a few-shot learning approach \citep{NEURIPS2020_1457c0d6, shi2022languagemodelsmultilingualchainofthought, ahuja-etal-2023-mega, fernandes2025deepseek}. Model families such as LLaMa \citep{grattafiori2024llama3}, Generative Pre-trained Transformer (GPT) \citep{openai2024gpt4technicalreport}, Qwen \citep{qwen2}, Claude \citep{anthropic2025}, Mistral \citep{jiang2023mistral7b, mistral_large_instruct_2407} or Gemma \citep{gemma2024, gemma2}, to name but a few, are experiencing continuous growth, offering LLMs with different parameter sizes and evaluated on various downstream tasks.

Nevertheless, there seems to be a noticeable difference in the performance of such models depending on the resources of the language and its complexity. \citet{arnett-bergen-2025-language} demonstrate that there is a disparity in the performance of LLMs between agglutinative and fusional languages, giving an advantage to high-resource languages like English over more morphologically complex languages such as Turkish. This is attributed to tokenization quality, stating that morphological alignment does not influence the model's performance. In one of the first systematic analyses of the morphological capabilities of LLMs for the tasks of inflection and reinflection, models such as ChatGPT are still far from achieving state-of-the-art results, performing on the level of some older supervised models \cite{weissweiler-etal-2023-counting}. 
In their multilingual version of the Wug test, \citet{anh-etal-2024-morphology} indicate that LLMs can apply their morphological knowledge to previously unseen words and that the morphological complexity of languages is more important than their relative representation in the training data. This fact states the importance of morphology for improving low-resource language modeling. 

Furthermore, it has also been demonstrated that LLMs could be used for lemma disambiguation in a dictionary-augmented approach for the endangered languages such as Erzya and Skolt Sami \citep{hämäläinen2024dag}, reaching the level close to a human annotator. 
Nonetheless, to the best of our knowledge, no empirical work has studied the ability of LLMs to generate correct lemmas.

The lack of quality annotated corpora for many target languages led to the exploration of model-transfer and data-transfer techniques. Model-transfer is based on the cross-lingual capabilities of multilingual pre-trained MLMs, where the knowledge in the source data can be transferred to the target language \citep{wang2023pretrained}.
Data-transfer aims to automatically produce labeled data
for the target language, traditionally based on translation and annotation approaches \citep{fei-etal-2020-cross}. Both techniques have demonstrated competitive results in the tasks of cross-lingual transfer for various sequence labeling tasks \citep{garcia-ferrero-etal-2022-model, chen-etal-2023-frustratingly}, although lemmatization has not been studied from a cross-lingual transfer perspective.

\section{Materials and Methods}\label{sec:materials_methods}

In this Section, we describe the datasets and models we use for our experiments. We also describe the prompting method we apply to perform in-context lemma generation.

\subsection{Datasets}
\label{sec:datasets}
In order to address the three RQs, we use 2 types of corpora, described below.

To address RQ1 and RQ3, we use parallel corpora, namely, the PUD treebank, presented for the CoNLL 2017 Shared Task \citep{zeman-etal-2017-conll}. This corpus was created for 18 languages, and each PUD dataset consists of 1000 sentences extracted from online sources and Wikipedia; 750 of 1000 sentences were originally in English, while the rest came from German, Spanish, French, and Italian texts. The corpora were further translated by professional translators to the remaining languages. 

Regarding RQ1, we chose a limited scope of languages from our selection in order to test the in-context lemma generation using LLMs, namely, English, Spanish, Russian, and Basque. As Basque does not have parallel corpora in the PUD treebank, we took the first 100 sentences of the BDT test corpus to establish an equal experimental setup. Such language selection was motivated by our ability to conduct detailed in-house analysis of the obtained results, as well as to determine the optimal model settings.

Concerning the experimentation to answer RQ3, we chose 12 languages of different morphological complexity, namely English, Spanish, French, German, Italian, Finnish, Icelandic, Turkish, Swedish, Czech, Polish, and Russian. In order to perform model transfer experiments, we split the PUD data into standard training, development, and test partitions, resulting in 800 sentences for training and 100 sentences for the development and test sets, respectively. 

To respond to RQ2, we use datasets of different data distributions for fine-tuning and testing in Basque, Czech, English, Spanish, Russian, and Turkish. The datasets were developed for the SIGMORPHON 2019 Shared Task on contextual lemmatization \citep{mccarthy-etal-2019-sigmorphon}. The data is annotated according to the Unimorph scheme \citep{mccarthy-etal-2020-unimorph}, the only exception being the Basque Armiarma corpus, an external dataset to UD, which includes lemma annotations of literary critics \citep{armiirma2000}. This selection of languages and datasets allows us to compare with previous out-of-domain lemmatization results \cite{toporkov-agerri-2024-role}. In order to make the computational load more manageable, the number of sentences in the larger datasets is reduced to 900. Statistics regarding the token and sentence counts in the final sets are presented in the Table \ref{tab:ood_corpora}.

 \begin{table}[ht]
 \small
\centering
\resizebox{\columnwidth}{!}{%
\begin{tabular}{llrr}
\toprule
\textbf{Language} & \textbf{Corpus} & \textbf{Tokens} & \textbf{Sentences} \\
\midrule
Basque   & BDT   & 11901 & 900 \\
Basque   & Armiarma  & 17172 & 900 \\
Spanish  & AnCora & 26917 & 900 \\
Spanish  & GSD & 24412 & 900 \\
English  & EWT & 13690 & 900 \\
English  & GUM  & 8189  & 440 \\
Turkish  & IMST  & 5734  & 564 \\
Turkish  & PUD  & 1795  & 100 \\
Czech    & CAC  & 17855 & 900 \\
Czech    & PUD & 1930  & 100 \\
Russian  & GSD  & 9874  & 503 \\
Russian  & SynTagRus & 16594 & 900 \\
\bottomrule
\end{tabular}%
}
\caption{Datasets for the experiments on corpora across different distributions.}
\label{tab:ood_corpora}
\end{table}

\subsection{Models}
\label{sec:models}

We evaluate the performance of several state-of-the-art multilingual instruction-tuned generative large language models, specifically, Mistral-Large-Instruct-2407 \citep{mistral_large_instruct_2407}, LLaMA-3.3-70B-Instruct \citep{grattafiori2024llama3},  Qwen-2.5-72B-Instruct \citep{qwen2}, and Claude-3.7-Sonnet \citep{anthropic2025} (100B+ paremeters). All models are evaluated using zero-shot and few-shot (1, 2, 3, 4 and 5) in-context learning. We use the default configuration for all LLMs. Every model except Claude has publicly released their weights.



For the contrastive supervised approach, we apply the large version of XLM-RoBERTa \citep{conneau-etal-2020-unsupervised}. This encoder model is based on the RoBERTa architecture and was pre-trained on 2.5TB of filtered CommonCrawl data for 100 languages and has obtained state-of-the-art results for many discriminative tasks \cite{garcia-ferrero-etal-2022-model}, such as cross-lingual sequence labeling.

\subsection{Prompt Design}\label{sec:prompt_design}
To establish which prompt could provide us with the optimal results in the lemma generation task, we try 4 different prompt settings to perform zero-shot and few-shot experiments. As mentioned earlier, we experiment with 4 languages of different morphological complexity, namely English, Russian, Spanish, and Basque. 

For each prompt type, we introduce two different inputs to the model: the whole sentence as one string and the sentence introduced as a list of words. We aim to receive the output as a whole sentence presented in the form of \emph{word-lemma} pairs. The instructions provided for the models are always given in English. The prompts are designed in the following fashion:
\begin{itemize}[noitemsep, topsep=0pt, label=--]
  \item \textit{basic prompt}: simple description of the task;
  \item \textit{full prompt}: simple description of the task + lemmatization instructions;
  \item \textit{basic prompt + k-shot examples [1:5]};
  \item \textit{full prompt + k-shot examples [1:5]}.
\end{itemize}

\textbf{\textit{Simple description of the task (basic prompt)}}. The basic prompt consists of a brief description of the lemmatization task without specifying any particular instructions. The example of such a prompt is presented below:

\begin{tcolorbox}[colback=white, colframe=black, boxrule=0.5pt]
\small\textit{"Your task is to lemmatize a sentence in Spanish. You will be given a sentence, where each word starts from the new line. You need to provide for each word in the given sentence its dictionary form (lemma).\\
Provide the output in **TSV format** (Tab-Separated Values) with the format: \\
`initial word \hspace{0.4cm} lemma' \\
Sentence: "El Parque Golden Gate ofrece un jardín botánico , un planetario , y un jardín japonés ." \\
Answer with the required output only, without extra spaces, quotation marks, or comments."}
\end{tcolorbox}

\textbf{\textit{Simple description of the task + explicit lemmatization instructions (full prompt)}}.
The preliminary output analysis using the basic prompt demonstrated that the model was skipping particular words, introducing amendments to the existing words (e.g., if the word was misspelled in the original corpus), or struggling with specific particles such as the articles in Spanish. Therefore, to improve the LLM's performance, we accompanied the prompt with a detailed set of instructions to perform the lemmatization task, presented below:

\begin{tcolorbox}[colback=white, colframe=black, boxrule=0.5pt]
\small\textit{Instructions:\\
            1. Copy the word exactly as it is, and provide its lemma.\\
            2. **Process Every Word**: Lemmatize **each word** in the sentence. Do not omit, change, or remove any word.\\
            3. **Handle Spelling Errors**: If a word is misspelled, retain the original spelling as the initial word, but lemmatize it to the closest dictionary form.\\
            4. **Proper Nouns**: Proper nouns should retain their capitalization.\\
            5. **Punctuation**: Include punctuation marks in the output, using the mark itself as the lemma.\\
            6. **Part-of-Speech**: Lemmatize words based on their part of speech (POS) (e.g., verbs to their infinitive form, nouns to singular form).\\
            7. **Articles**: Use the masculine singular form for articles.\\
            8. **Multi-Word Expressions**: If an input contains multiple words, process each word separately.\\}
\end{tcolorbox}

\textbf{\textit{Basic prompt and full prompt + k-shot examples}}.
The third and fourth prompts include examples. In order to choose the examples for the few-shot experiments, we experimented with the development set of the PUD corpus using Mistral-Large-Instruct-2407. We tried manual example selection, random example selection, and choosing the examples with the highest number of errors committed, such as skipping one of the words in the sentence, wrong lemma generation, and generation of additional elements. We then introduce examples in a range of 1 to 5 to each prompt configuration, namely, basic and full (an example of the prompt is given in Appendix \ref{prompt_example}). The prompt configuration with the best results on this particular experiment is then applied to the rest of our experiments.

\section{Experiments}
\label{sec:experiments}

In this paper, we aim to answer the following questions: (RQ1) What are the capabilities of the latest generation of LLMs to perform in-context learning for lemmatization? (RQ2) Can we use direct lemma generation to address the problem of poor out-of-domain performance, namely, when evaluated on a data distribution different from the one seen during training? (RQ3) What is the optimal strategy in a scenario in which annotated training data in the target domain or language is not available?

To address these questions, we conduct three sets of experiments. Each experimental setup and its corresponding results are described below in a separate subsection. Evaluation is performed using average word and sentence accuracy metrics.\footnote{Sentence accuracy allows for better discrimination between models' performance \cite{toporkov-agerri-2024-role}.} We conduct 3 runs and report the average results for all experiments and models, except for Claude-3.7-Sonnet, where only 1 run was performed due to computational costs. To assess whether the observed differences in model performance are statistically significant, we apply McNemar's test \citep{dietterich}.

\subsection{In-context Lemma Generation}

To respond to RQ1, we experiment with direct lemmatization using LLMs, employing Mistral-Large-Instruct-2407 and the prompt types described in Section \ref{sec:prompt_design}. This model was selected for its strong performance-to-cost ratio, fast inference times, and accessibility within our computational environment. 
For each prompt, 3 different runs are performed using the datasets described in Section \ref{sec:datasets} on English, Spanish, Russian, and Basque. The most effective prompt setting will be identified based on word and sentence accuracy results, performance on highly inflected languages such as Russian and Basque, and the number of hallucinations and errors.

\begin{table*}[t] 
\centering
\resizebox{\textwidth}{!}{%
\begin{tabular}{@{}lllrrrrrrrrl@{}}
\toprule
& & & \multicolumn{4}{c}{\textit{input: sentence as one string}}  & \multicolumn{4}{c}{\textit{input: sentence as a list of words}} \\  \cmidrule{4-11}
& & & {\textit{WAcc}} & {\textit{SentAcc}} &  {\textit{missing words}} & {\textit{wrong words}} &  {\textit{WAcc}} &  {\textit{SentAcc}} &  {\textit{missing words}} & {\textit{wrong words}} \\ \cmidrule{4-11}
\textit{en} & \textit{basic prompt} & \textit{0-shot} & 0.93 & 0.28 & 30 & 0 & 0.95  & 0.38 & 0 & 0 \\
 & \textit{full prompt}  & \textit{0-shot} & 0.95 & 0.43 & 4 & 0 & 0.95 & 0.42 & 2 & 0 \\
  & \textit{basic prompt} & \textit{4-shot} & 0.96 & 0.42 & 0 & 0 & 0.96 & 0.44 & 0 & 0 \\
   & \textit{full prompt} & \textit{4-shot} & \textbf{0.96} & \textbf{0.46} & 0 & 0 & 0.95 & 0.43 & 0 & 0 \\ \hline
\textit{es} & \textit{basic prompt} & \textit{0-shot} & 0.92 & 0.12  & 32 & 2 & 0.93  & 0.22 & 0 & 0 \\
  & \textit{full prompt}  & \textit{0-shot} & 0.93 & 0.22 & 4 & 3 & 0.92 & 0.17 & 1 & 0 \\
  & \textit{basic prompt} & \textit{4-shot} & 0.96 & 0.41 & 2 & 0 & \textbf{0.97} & \textbf{0.45} & 1 & 0 \\
   & \textit{full prompt} & \textit{4-shot} & 0.96 & 0.42 & 2 & 0 & 0.96 & 0.43 & 0  & 0 \\ \hline
\textit{ru} & \textit{basic prompt} & \textit{0-shot} & 0.91 & 0.20 & 32 & 5 & 0.92 & 0.29 & 0 & 2 \\
 & \textit{full prompt} & \textit{0-shot} & 0.94 & 0.33 & 4 & 3 & 0.94 & 0.34  & 0 & 4 \\
& \textit{basic prompt} & \textit{4-shot} & 0.93 & 0.41 & 42  & 0  & \textbf{0.95}  & \textbf{0.40}  & 2 & 0 \\
& \textit{full prompt} & \textit{4-shot} & 0.93 & 0.37 & 42 & 0  & 0.94  & 0.37 & 1 & 0 \\ \hline
\textit{eu} & \textit{basic prompt} & \textit{0-shot} & 0.76  & 0.00 & 78 & 3 & 0.82 & 0.07  & 3 & 0 \\
 & \textit{full prompt} & \textit{0-shot} & 0.78 & 0.03 & 44 & 3 & 0.81  & 0.05 & 1  & 0 \\
& \textit{basic prompt} & \textit{4-shot} & 0.86  & 0.23 & 38  & 0 & \textbf{0.89} & \textbf{0.29} & 7  & 0 \\
& \textit{full prompt} & \textit{4-shot} & 0.86 & 0.25& 34 & 0  & 0.89  & 0.27  & 1 & 0 \\ \bottomrule
\end{tabular}
}
\caption{Word and sentence accuracy using different prompting strategies with Mistral-Large-Instruct-2407. In \textbf{bold}: best overall accuracy results per language.}
\label{tab:results_prompting}
\end{table*}

Table \ref{tab:results_prompting} reports the results on 0-shot and 4-shot (best overall configuration in terms of the number of examples) experiments combining different prompt strategies. It can be observed that 
when given the basic prompt
without specific instructions and in a zero-shot setting, the model fails to generate a lot of input words across all languages.

Adding some examples to the prompt significantly improves the results, especially for more complex languages such as Russian and Basque. The input format also plays a role: with the sentence represented as separated tokens, the model hallucinates less and obtains higher accuracies. As stated by \citet{arnett-bergen-2025-language}, this could be directly connected to the tokenization quality. Surprisingly, introducing example selection (described in Section \ref{sec:prompt_design}) into the prompt makes the basic prompt outperform the full one when the input is a list of tokens.

In addition to word and sentence accuracy, the optimal prompt design is chosen based on the amount of produced errors, hallucinations, and missing words. Thus, Table \ref{tab:results_prompting} shows that the best setting corresponds to the \textbf{\textit{basic prompt + 4-shot examples}}, where the input is a sentence as a list of words, and the example selection is based on ranking sentences by the number of errors the model produced during the inference on the development set. The explicit word and sentence accuracy results for each of the prompt types are detailed in Appendix \ref{sec:appendixA}.

\subsection{Experiments with Corpora of Different Distributions}

In the second part of our experiments, we aim to determine the best strategy for performing the lemmatization task on data from a distribution different from the one seen during training. We take the experiments by \citet{toporkov-agerri-2024-role} as a starting point, where they show that lemmatizers (based on fine-tuned encoder-only models) substantially worsen when evaluated out-of-domain, their most common use case. Therefore, in this setting, we compare the performance of encoder models fine-tuned on gold-annotated corpora and applied out-of-domain against the direct lemma generation based on in-context learning using LLMs. For this purpose, we employ datasets and models described in Section \ref{sec:materials_methods}. We fine-tune XLM-RoBERTa large for each of the 6 languages of our selection in a token classification task, where the model learns to predict labels corresponding to the minimum number of edits required to transform the target word into its lemma. We use 16 as a batch size, 0.01 weight decay, 5e-5 learning rate, and 20 epochs as hyperparameters. For the in-context lemma generation, we choose the best prompt setting from the previous experimental step and apply it using the models described in Section \ref{sec:models}.

\begin{table*}[t]

\centering
\resizebox{\textwidth}{!}{%
\begin{tabular}{l|cc|cc|cc|cc||cc}
\toprule
\multicolumn{1}{l}{}  & \multicolumn{2}{c}{\textbf{Mistral-LI-2407}} & \multicolumn{2}{c}{\textbf{Llama-3.3-70B}} & \multicolumn{2}{c}{\textbf{Qwen-2.5-72B}} & \multicolumn{2}{c}{\textbf{Claude-3.7}}  & \multicolumn{2}{c}{\textbf{XLM-R large}} \\ 
\textbf{language\_corpus} & {\textit{Wacc}} & {\textit{SentAcc}} &{\textit{Wacc}} & {\textit{SentAcc}} & {\textit{Wacc}} & {\textit{SentAcc}} &{\textit{Wacc}} & {\textit{SentAcc}} & {\textit{Wacc}} & {\textit{SentAcc}} \\ \hline
eu\_bdt \footnotemark & 0.84  & 0.12 & 0.78 & 0.05 & 0.71 & 0.02 & \textbf{0.89} & \textbf{0.20} & {-} & {-} \\
eu\_armiarma & 0.83 & 0.08 & 0.75 & 0.04 & 0.70 & 0.02 & 0.88 & 0.15 & \textbf{0.89}\textsuperscript{*} & \textbf{0.18}\textsuperscript{*} \\
es\_ancora  & 0.93 & 0.23 & 0.87 & 0.16 & 0.92 & 0.24 & 0.94 & 0.32 & \textbf{0.97}\textsuperscript{*}  & \textbf{0.51}\textsuperscript{*}  \\
es\_gsd & 0.93 & 0.25 & 0.87 & 0.15 & 0.91 & 0.21 & 0.94 & 0.32 & \textbf{0.96}\textsuperscript{*} & \textbf{0.43}\textsuperscript{*} \\
en\_ewt & \textbf{0.93}\textsuperscript{*} & \textbf{0.44}\textsuperscript{*} & 0.92 & 0.39 & 0.92 & 0.42 & 0.93 & 0.33 & 0.92 & 0.39   \\
en\_gum  &\textbf{ 0.94} & \textbf{0.47} & 0.92 & 0.40 & 0.93 & 0.46 & 0.94 & 0.37 & 0.94  & 0.43 \\
tr\_imst   & 0.90  & 0.44 & 0.86 & 0.31 & 0.84 & 0.31 & \textbf{0.94}\textsuperscript{*}  & \textbf{0.60}\textsuperscript{*} & 0.81 & 0.19  \\
tr\_pud (sigm'19) & 0.81 & 0.06 & 0.80 & 0.04 & 0.77 & 0.01 & 0.84 & 0.08 & \textbf{0.85}  & \textbf{0.08} \\
cs\_cac & 0.94 & 0.39   & 0.87 & 0.21 & 0.89 & 0.19 & \textbf{0.97}\textsuperscript{*} & \textbf{0.55}\textsuperscript{*} & 0.93 & 0.32  \\
cs\_pud (sigm'19) & 0.95 & 0.34 & 0.91 & 0.24 & 0.89 & 0.15 & \textbf{0.97}\textsuperscript{*}  & \textbf{0.55}\textsuperscript{*}  & 0.95 & 0.43  \\
ru\_gsd & 0.94 & 0.41& 0.87 & 0.26 & 0.92 & 0.29 & \textbf{0.96}\textsuperscript{*} & \textbf{0.51}\textsuperscript{*} & 0.94  & 0.39 \\
ru\_syntagrus & 0.95 & 0.43 & 0.91 & 0.34 & 0.93 & 0.35 & \textbf{0.96}\textsuperscript{*} & \textbf{0.48}\textsuperscript{*} & 0.94 & 0.41  \\
\bottomrule
\textit{average} & 0.91 & 0.32 & 0.87 & 0.23 & 0.87 & 0.24 & \textbf{0.93} & \textbf{0.39}   & 0.92 & 0.34 \\ \bottomrule
\end{tabular}
}
\caption{Word and sentence accuracy results comparing in-context lemma generation against XLM-RoBERTa large performance on lemmatization fine-tuned on a different data distribution. In \textbf{bold}: best overall accuracy results across models for each language. *:statistically significant results at $\alpha = .05$.}
\label{tab:results_ood}
\end{table*}

Table \ref{tab:results_ood} reports the results.
We mark in bold the best overall result for each language and with an asterisk those results that are statistically significant according to McNemar's test.
As previously mentioned in Section \ref{sec:experiments}, each result corresponds to the average over 3 runs. The standard deviation for both word and sentence accuracy across these runs is consistently low ($\leq 0.02$), indicating stable behavior of the models.

We could see that directly generating the lemmas with \emph{Claude} and \emph{Mistral} outperforms the fine-tuned encoder in Turkish, Czech, and Russian. For English, the results across models are very close, although they are only statistically significant for EWT corpus. For Basque and Spanish XLM-RoBERTa large evaluated out-of-domain is the superior option. Among LLMs, the highest accuracy for 6 out of 12 corpora is achieved using Claude-3.7-Sonnet, while Mistral-Large-Instruct-2407 ranks a close second for Spanish, English, and Russian. It is worth noticing that the open-weights Mistral significantly outperforms XLM-RoBERTA large in 6 of the 11 evaluation settings.

\subsection{Experiments with Parallel Corpora}

To address RQ3, we perform a set of experiments using cross-lingual transfer and in-context lemma generation. For both model- and data-transfer, we assume our source language to be English. For \emph{model-transfer}, we train a contextual lemmatizer on English data using XLM-RoBERTa large in a token classification task, applying the same hyperparameters as in the experiments of the previous section. The \emph{data-transfer} method, implemented as translate-train, requires the automatic generation of the training data for each of the target languages. We apply Claude-3.7-Sonnet to translate the PUD training set to the 11 target languages and to obtain the lemmas directly from the translations. We then fine-tune XLM-RoBERTa large using the generated training set.

\begin{table*}[t]
\centering
\resizebox{\textwidth}{!}{%
\begin{tabular}{l|cc|cc|cc|cc||cc|cc|cc}
\toprule
\multicolumn{1}{l}{} & \multicolumn{2}{c}{\textbf{Mistral-LI-2407}} & \multicolumn{2}{c}{\textbf{Llama-3.3-70B}} & \multicolumn{2}{c}{\textbf{Qwen-2.5-72B}} & \multicolumn{2}{c}{\textbf{Claude-3-7}}  & \multicolumn{2}{c}{\textbf{monolingual}} & \multicolumn{2}{c}{\textbf{model-transfer}} & \multicolumn{2}{c}{\textbf{data-transfer}}  \\ \hline
\multicolumn{1}{l}{\textbf{language}} & {\textit{Wacc}} & {\textit{SentAcc}} & {\textit{Wacc}} & {\textit{SentAcc}} & {\textit{Wacc}} & {\textit{SentAcc}} & {\textit{Wacc}} & {\textit{SentAcc}} & {\textit{Wacc}} & {\textit{SentAcc}} & {\textit{Wacc}} & {\textit{SentAcc}} &{\textit{Wacc}} & {\textit{SentAcc}} \\ \hline
en  & \underline{0.96}  & \underline{0.44} & 0.94   & 0.35   & 0.95    & 0.42  & 0.94  & 0.30  & \textbf{0.97}   & \textbf{0.58}  & -  & -  & -  & - \\
de  & 0.96  & 0.51 & 0.93 & 0.25 & 0.93 & 0.31 & \underline{\textbf{0.97}} & \underline{\textbf{0.64}}  & \underline{0.95} & 0.37 & 0.64 & 0.00 & \underline{0.95} & \underline{0.39} \\
is  & 0.90  & 0.13  & 0.84  & 0.10  & 0.80 & 0.01 & \underline{\textbf{0.94}} & \underline{\textbf{0.30}}  & \underline{0.89} & \underline{0.09} & 0.73 & 0.00 & 0.87 & 0.03  \\
sv & 0.94 & 0.34 & 0.91 & 0.21 & 0.88 & 0.12 & \underline{0.95} & \underline{0.36} & \underline{\textbf{0.96}} & \underline{\textbf{0.48}} & 0.76 & 0.00  & 0.92 & 0.26   \\
ru  & 0.95 & 0.40  & 0.93 & 0.36 & 0.93 & 0.30 & \underline{\textbf{0.96}} & \underline{\textbf{0.49}} & \underline{0.94} & \underline{0.27} & 0.76 & 0.00 & 0.92 & 0.21  \\
cs  & 0.95 & 0.37 & 0.91  & 0.23  & 0.95 & 0.24 & \underline{\textbf{0.97}} & \underline{\textbf{0.65}} & \underline{0.95} & \underline{0.45} & 0.80 & 0.00 & 0.94 & 0.35  \\
pl & 0.95 & 0.42 & 0.92 & 0.22 & 0.90 & 0.17 & \underline{\textbf{0.95}} & \underline{\textbf{0.46}} & \underline{0.94} & \underline{0.36} & 0.75 & 0.00 & 0.91 & 0.18   \\
es & 0.97 & 0.45 & 0.94 & 0.31  & 0.96 & 0.40 & \underline{\textbf{0.98}} & \underline{\textbf{0.65}} & \underline{0.97} & \underline{0.55} & 0.78 & 0.00 & 0.96 & 0.38  \\
it & 0.96 & 0.37 & 0.92 & 0.25 & 0.94 & 0.26 & \underline{0.97} & \underline{0.44} & \underline{\textbf{0.97}} & \underline{\textbf{0.55}} & 0.78 & 0.00 & 0.95 & 0.34 \\
fr & 0.97 & \underline{0.51} & 0.94 & 0.26 & 0.95 & 0.24 & \underline{0.98} & 0.42 & \underline{\textbf{0.98}} & \underline{\textbf{0.59}} & 0.78 & 0.00 & 0.96 & 0.36  \\
fi & 0.89 & 0.15 & 0.84 & 0.09 & 0.80 & 0.03 & \underline{\textbf{0.93}} & \underline{\textbf{0.35}}  & \underline{0.86}  & \underline{0.06}  & 0.77 & 0.00 & 0.82 & 0.05 \\
tr & 0.81 & 0.03 & 0.79  & 0.02 & 0.78 & 0.01 & \underline{0.85} & \underline{0.06} & \underline{\textbf{0.88}} & \underline{\textbf{0.16}} & 0.75 & 0.00  & 0.82  & 0.03  \\ 
\bottomrule
\textit{average} & 0.93 & 0.34** & 0.90  & 0.22  & 0.90 & 0.21 & \underline{\textbf{0.95}} & \underline{\textbf{0.43}} & \underline{\textbf{0.94}} & \underline{\textbf{0.38}}  & 0.75  & 0.00 & 0.91  & 0.23 \\ \bottomrule
\end{tabular}
}
\captionof{table}{Word and sentence accuracy across various approaches to lemmatization task using parallel corpora. In \textbf{bold}: best overall accuracy per language. \underline{Underlined}: best model for the language for each subset of models.}
\label{tab:results_pud}
\end{table*}

Table \ref{tab:results_pud} displays the results of the experiments conducted with the parallel corpora. We mark in bold the best overall result per language and underline the best model for the language per each subset of models (decoder-only vs. encoder-only models). We observe that the latest generation of LLMs demonstrates strong results in the lemma generation task, and the models' performance remains stable across consecutive runs, as in the previous experimental setup (standard deviation $\leq 0.02$). It should be emphasized that we explore the performance of these models using in-context learning, without them being specifically fine-tuned for lemmatization. Among the 4 models of our choice, Claude-3.7-Sonnet exhibits the highest accuracy for all the languages except English, outperforming Mistral-Large-Instruct-2407, LLaMA-3.3-70B-Instruct, and Qwen-2.5-72B-Instruct. Still, the open-weights Mistral-Large-Instruct-2407 stays a close second in this setting, while the other two models demonstrate lower capabilities for the task.

Cross-lingual transfer obtains less competitive results, suggesting that, in data scarcity scenarios, directly generating lemmas with LLMs is a very effective method, far superior to using traditional model-transfer or data-transfer approaches (at least for the set of languages considered).  

Surprisingly, direct lemma generation using \emph{Claude} or \emph{Mistral} is superior to even fine-tuning XLM-RoBERTA large in-domain (e.g., monolingual results) for 7 out of the 12 languages, although for English, Swedish, Italian, French, and Turkish XLM-RoBERTA remains the best option. It can also be observed that significant differences in word and sentence accuracy are demonstrated only for Turkish, Swedish, and English, while for French and Italian the distinction could be perceived only by comparing sentence accuracy results. This highlights the importance of using sentence accuracy as an alternative metric to word accuracy \cite{Manning2011PartofSpeechTF,toporkov-agerri-2024-role}.

Finally, results obtained with the data-transfer approach exceed those of direct in-context lemma generation using \textit{LLaMa} or \textit{Qwen} for 6 out of 11 languages, indicating the high quality of translations produced by \textit{Claude}.

To perform all the experiments we utilized several NVIDIA A100 80GB GPUs and the total computational resources amounted to approximately 300 hours of processing time and 75 kWh of energy consumption, resulting in 32.4 kg of CO2 emissions.

\footnotetext{
Since the Armiarma dataset lacks training data, we could not perform out-of-domain experiments on the BDT test set, and thus results for XLM-R large model are not reported.}

\section{Discussion}
\label{sec:discussion}

Although LLMs show promising results in in-context lemma generation for the selected languages, their performance depends heavily on the provided examples. In prompt development across 4 languages of varied morphological complexity, the differences between the basic in a zero-shot setting and the 4-shot prompt were quite significant. We identified a number of problems when using LLMs for direct lemma generation, such as:
(i) \textit{randomly generated output.} LLMs such as Mistral-Large-Instruct-2407 and Claude-3.7-Sonnet tend to randomly generate quotation marks, but there are also cases, exhibited for English (EWT corpus), where Claude-3.7-Sonnet provided the output for the same sentence more than once. In the case of LLaMA-3.3-70B-Instruct, the model gives additional explanations, for instance, in the case of ambiguous Basque auxiliary verbs such as \textit{edun} and \textit{izan}.
(ii) \textit{modified wordforms.} Despite the explicit instructions on 
not performing any changes to the initial word, even if it is misspelled, LLMs tend to ignore them (this was the case during the experimental prompting phase with Mistral-Large-Instruct-2407). Apart from that, it is common to lowercase the input words at the beginning of the sentence. 
(iii) \textit{arbitrarily skipping words.} This was the most interesting observation, as there was no perceivable pattern of which words the model was skipping. For instance, for Basque Mistral-Large-Instruct-2407 ignored verbal forms such as `egingo' (\textit{`will do'} in English); 
(iv) \textit{struggling with certain lemmas that do not appear in the few-shot examples}. This case was observed for articles in Spanish for which, instead of providing the lemma for definite articles such as `\textit{el}', LLMs were returning the same word form of the determiner given in the input text (e.g., \textit{la}, \textit{los} or \textit{las}). This behaviour could be changed by adding an example in the prompt to deal with these particular cases.

Model scale also plays a crucial role in the quality of in-context lemmatization. Table \ref{tab:results_different_scales_pud} presents comparative results for the smaller and larger versions of the \textit{Qwen} and \textit{Mistral} models, 
indicating that larger models demonstrate significantly stronger performance on the lemmatization task. Additionally, we experimented with LLaMa-3.1-8B, but the results were difficult to report, as the model failed to generate a coherent output for 10 out of 12 languages, performing a lot of hallucinations, incorrect output format and numerous sentence repetitions.

\begin{table*}[t]
\centering
\resizebox{\textwidth}{!}{%
\begin{tabular}{l|cc|cc|cc|cc|cc}
\toprule
\multicolumn{1}{l}{} & \multicolumn{2}{c}{\textbf{Qwen-2.5-7B}} & \multicolumn{2}{c}{\textbf{Qwen-2.5-32B}} & \multicolumn{2}{c}{\textbf{Qwen-2.5-72B}} & \multicolumn{2}{c}{\textbf{Ministral-8B}} & \multicolumn{2}{c}{\textbf{Mistral-LI-2407}} \\ 
\multicolumn{1}{l}{\textbf{language}} & \multicolumn{1}{c}{\textit{Wacc}} & \multicolumn{1}{c}{\textit{SentAcc}} & \multicolumn{1}{c}{\textit{Wacc}} & \multicolumn{1}{c}{\textit{SentAcc}} & \multicolumn{1}{c}{\textit{Wacc}} & \multicolumn{1}{c}{\textit{SentAcc}} & \multicolumn{1}{c}{\textit{Wacc}} & \multicolumn{1}{c}{\textit{SentAcc}} & \multicolumn{1}{c}{\textit{Wacc}} & \multicolumn{1}{c}{\textit{SentAcc}} \\ 
\midrule
en & 0.81 & 0.08 & 0.90 & 0.26 & 0.95 & 0.42 & 0.85 & 0.09 & 0.96 & 0.44 \\
de & 0.72 & 0.00 & 0.90 & 0.16 & 0.93 & 0.31 & 0.77 & 0.01 & 0.96 & 0.51 \\
is  & 0.62 & 0.00 & 0.77 & 0.01 & 0.80 & 0.01 & 0.71 & 0.01 & 0.90 & 0.13 \\
sv & 0.75 & 0.00 & 0.86 & 0.06 & 0.88 & 0.12 & 0.80 & 0.02 & 0.94 & 0.34 \\
ru  & 0.89 & 0.13 & 0.93 & 0.25 & 0.93 & 0.30 & 0.82 & 0.16 & 0.95 & 0.40 \\
cs & 0.78 & 0.03 & 0.87 & 0.07 & 0.95 & 0.24 & 0.81 & 0.03 & 0.95 & 0.37 \\
pl & 0.75 & 0.02 & 0.88 & 0.11 & 0.90 & 0.17 & 0.80 & 0.00 & 0.95 & 0.42 \\
es & 0.78 & 0.01 & 0.93 & 0.23 & 0.96 & 0.40 & 0.81 & 0.06 & 0.97 & 0.45 \\
it & 0.76 & 0.00 & 0.92 & 0.17 & 0.94 & 0.26 & 0.82 & 0.04 & 0.96 & 0.37 \\
fr & 0.70 & 0.00 & 0.92 & 0.19 & 0.95 & 0.24 & 0.83 & 0.03 & 0.97 & 0.51 \\
fi  & 0.65 & 0.00 & 0.78 & 0.03 & 0.80 & 0.03 & 0.72 & 0.01 & 0.89 & 0.15 \\
tr & 0.66 & 0.00 & 0.76 & 0.03 & 0.78 & 0.01 & 0.71 & 0.00 & 0.81 & 0.03 \\
\bottomrule
\textit{average} & 0.74 & 0.02 & 0.87 & 0.13 & 0.90 & 0.21 & 0.79 & 0.04 & 0.93 & 0.34 \\ \bottomrule
\end{tabular}
}
\captionof{table}{Word and sentence accuracy for in-context lemma generation across different model sizes using parallel corpora.}
\label{tab:results_different_scales_pud}
\end{table*}

These are only some of the most common observations obtained during the analysis of LLMs' performance on contextual lemmatization. The quality of examples used in the prompt design plays an important role, and careful and elaborated example selection may be specifically beneficial for low-resource languages. A deeper analysis should be conducted regarding other potential pitfalls of LLMs for this task, especially for languages with more complex morphology.

\section{Conclusions}

In this paper, we present the first empirical analysis of the ability of the latest-generation LLMs to perform in-context lemma generation. Our results suggest that, although fine-tuning encoders such as XLM-RoBERTa large on gold data remains a competitive option for its use out-of-domain, large size LLMs still reach results close to the state-of-the-art by directly generating lemmas in-context in a few-shot setting. We also investigate the scenario in which no training data is available for a given language. Comparing model-transfer, data-transfer, and direct lemma generation with LLMs, we conclude that the best lemmatization approach in such a case would also be direct in-context lemma generation, which would remain predominantly achievable with large-scale language models. Finally, future work should include model comparison, broader linguistic sampling, and comprehensive prompt optimization.

Lemmatization is a task generally studied and evaluated in-domain, which is rather surprising as the use of lemmatizers is predominantly out-of-domain \cite{toporkov-agerri-2024-role}. Our work demonstrates, for the first time, the potential to perform direct lemmatization directly without any training data by applying direct in-context learning with large size LLMs, even for high-inflected relatively low-resource languages.




\section*{Limitations}

Several limitations constrain this investigation. First, the comprehensive evaluation across the full spectrum of large-scale language models remains unexplored, potentially excluding architectures that may demonstrate superior lemmatization capabilities. Second, the scarcity of available evaluation datasets, particularly for low-resource and morphologically complex languages, limits our ability to conduct extensive cross-linguistic validation. Third, we did not systematically explore alternative prompt variations or instructional formats.

\section*{Acknowledgments}

This work has been partially supported by the Basque Government (Research group funding IT-1805-22). We are also thankful to the following MCIN/AEI/10.13039/501100011033 projects: (i) DeepKnowledge (PID2021-127777OB-C21) and by FEDER, EU; (ii) DeepMinor (CNS2023-144375)
and European Union NextGenerationEU/PRTR.

\bibliography{anthology,bibliography}

\newpage
\appendix

\section{Appendix}
\label{sec:appendixA}

\begin{table*}[ht]
\centering
\resizebox{\textwidth}{!}{%

\begin{tabular}{@{}lllrrrrrrrrrr@{}}
\toprule
& & & \multicolumn{5}{c}{Sentence input} & \multicolumn{5}{c}{Wordform input} \\ \midrule
Language & Prompt Type  & Shots & WAcc & SentAcc & Missing & Wrong word & Random & WAcc & SentAcc & Missing & Wrong word & Random\\
\midrule

\multirow{6}{*}{\centering{Spanish (PUD)}} & basic prompt & 0-shot  & 0.92  & 0.12  & 32  & 2 & 2 & 0.93  & 0.22 & 0  & 0 & 9 \\
                             & full prompt  & 0-shot  & 0.93 & 0.22 & 4 & 3 & 6 & 0.92 & 0.17 & 1 & 0 & 6 \\
                             & basic prompt + worst examples & 4-shot & 0.96 & 0.37 & 3 & 0 & 3 & 0.96 & 0.38 & 1 & 0 & 0 \\
                             & basic prompt + random examples & 4-shot & 0.97 & 0.50 & 5 & 0 & 0 & 0.94 & 0.33 & 0 & 0 & 0 \\
                             & basic prompt + most errors & 4-shot & 0.96 & 0.41 & 2 & 0 & 0 & 0.97 & 0.45 & 1 & 0 & 0 \\
                             & full prompt + worst examples & 4-shot & 0.97 & 0.41 & 3 & 0 & 3 & 0.97 & 0.40 & 0 & 0 & 0 \\
                             & full prompt + random examples & 4-shot & 0.96 & 0.49 & 1 & 0 & 0 & 0.95 & 0.41 & 0 & 0 & 0\\ 
                             & full prompt + most errors & 4-shot & 0.96 & 0.49 & 1 & 0 & 0 & 0.95 & 0.41 & 0 & 0 & 0\\ \midrule
\multirow{6}{*}{\centering{English (PUD)}} & basic prompt & 0-shot  & 0.93  & 0.28  & 30  & 0 & 0 & 0.95 & 0.38 & 0 & 0 & 6 \\
                             & full prompt  & 0-shot  & 0.95  & 0.43 & 4 & 0 & 1 & 0.95 & 0.42 & 2 & 0 & 12 \\
                             & basic prompt + worst examples & 4-shot & 0.96 & 0.50 & 0 & 0 & 0 & 0.95 & 0.36 & 0 & 0 & 0 \\
                             & basic prompt + random examples & 4-shot & 0.95 & 0.47 & 0 & 0 & 0 & 0.96 & 0.45 & 0 & 0 & 0\\
                             & basic prompt + most errors & 4-shot & 0.96 & 0.42 & 0 & 0 & 0 & 0.96 & 0.44 & 0 & 0 & 4 \\
                             & full prompt + worst examples & 4-shot & 0.96 & 0.46 & 0 & 0 & 0 & 0.95 & 0.41 & 0 & 0 & 0 \\
                             & full prompt + random examples & 4-shot & 0.96 & 0.50 & 0 & 0 & 0 & 0.96 & 0.42 & 0 & 0 & 0 \\
                             & full prompt + most errors & 4-shot & 0.96 & 0.46 & 0 & 0 & 0 & 0.95 & 0.43 & 0 & 0 & 0 \\ \midrule
\multirow{6}{*}{\centering{Russian (PUD)}} & basic prompt & 0-shot  & 0.91  & 0.20  & 32 & 5 & 8 & 0.92 & 0.29 & 0 & 2 & 6 \\
                             & full prompt  & 0-shot  & 0.94  & 0.33 & 4 & 3 & 3 & 0.94 & 0.34 & 0 & 4 & 4 \\
                             & basic prompt + worst examples & 4-shot & 0.94 & 0.39 & 3 & 0 & 0 & 0.95 & 0.39 & 2 & 0 & 0 \\
                             & basic prompt + random examples & 4-shot & 0.94 & 0.39 & 4 & 0 & 1 & 0.95 & 0.40 & 0 & 0 & 0\\
                             & basic prompt + most errors & 4-shot & 0.93 & 0.41 & 42 & 0 & 6 & 0.95 & 0.40 & 2 & 0 & 0 \\
                             & full prompt + worst examples & 4-shot & 0.94 & 0.40 & 3 & 0 & 0 & 0.95 & 0.36 & 2 & 0 & 0 \\
                             & full prompt + random examples & 4-shot & 0.94 & 0.38 & 5 & 0 & 1 & 0.95 & 0.39 & 2 & 0 & 0 \\ 
                             & full prompt + most errors & 4-shot & 0.93 & 0.37 & 42 & 0 & 0 & 0.94 & 0.37 & 1 & 0 & 0 \\ \midrule

\multirow{6}{*}{\centering{Basque (BDT, 100 sentences)}} & basic prompt & 0-shot  & 0.76  & 0.00  & 78  & 3 & 58 & 0.82 & 0.07 & 3 & 0 & 72 \\
                             & full prompt  & 0-shot  & 0.78 & 0.03 & 44 & 3 & 44 & 0.81 & 0.05 & 1 & 0 & 68 \\
                             & basic prompt + worst wacc & 4-shot & 0.85 & 0.11 & 23 & 0 & 2 & 0.89 & 0.24 & 0 & 0 & 2 \\
                             & basic prompt + random examples & 4-shot & 0.87 & 0.19 & 38 & 0 & 2 & 0.89 & 0.24 & 0 & 0 & 2 \\
                             & basic prompt + most errors & 4-shot & 0.86 & 0.23 & 38 & 0 &  2 & 0.89 & 0.29 & 7& 0 & 2\\
                             & full prompt + worst examples & 4-shot & 0.84 & 0.12 & 20 & 0 & 2 & 0.89 & 0.25 & 1 & 0 & 2\\
                             & full prompt + random examples & 4-shot & 0.87 & 0.22 & 39 & 0 & 2 & 0.88 & 0.22 & 0 & 0 & 2\\
                             & full prompt + most errors & 4-shot & 0.86 & 0.25 & 34 & 0 & 0 & 0.89 & 0.27 & 1 & 0 & 2 \\

\bottomrule
\end{tabular}
}
\caption{Word and sentence accuracy using different prompting strategies with Mistral-Large-Instruct-2407.}
\label{tab:prompting-selection}
\end{table*}


\begin{tcolorbox}[colback=white, colframe=gray, boxrule=0.5pt, title={Example of the full prompt with a 1-shot example, where the input sentence is introduced as a list of words.}]
\small\textit{Your task is to lemmatize a sentence in Spanish. You will be given a sentence, where each word starts from the new line. You need to provide for each word in the given sentence its dictionary form (lemma). For example, for the sentence:\\
Tina  \\
Anselmi \\ 
se \\
ocupo \\
sobre \\
todo \\
de \\ 
los \\
derechos \\
de \\
los \\ 
trabajadores \\ 
textiles \\
y \\
los \\
profesores \\
. \\
The desired output is: \\
Tina Tina \\
Anselmi Anselmi \\ 
se el \\
ocupó ocupar \\ 
sobre sobre \\ 
todo todo \\
de de \\
los el \\
derechos derecho \\
de de \\
los el \\
trabajadores trabajador \\ 
textiles textil \\
y y \\
los el \\
profesores profesor \\
. . \\
Provide the output in **TSV format** (Tab-Separated Values) with the format: `initial word \hspace{0.4cm} lemma' \\
Sentence: \\
\mbox{[}`El', `festival', `de', `Venecia', `cerró', `hoy', \\ `con', `la', `entrega', `de', `los', `premios', `que', \\ `coronaron', `a', `el', `realizador', `Alexander', \\
`Sokurov', `y', `a', `el' `actor', `Michael', \\ 
`Fassbender', `.' \mbox{]} \\
Answer with the required output only, without extra spaces, quotation marks, or comments.}
\label{prompt_example}
\end{tcolorbox}

\end{document}